%% file: main.tex
\documentclass[11pt]{article}
\usepackage{graphicx}
\usepackage{hyperref}
\usepackage{booktabs}
\usepackage{geometry}
\usepackage{subcaption}
\usepackage{adjustbox}
\usepackage{authblk}
\usepackage[utf8]{inputenc}
\usepackage[backend=biber,style=numeric]{biblatex}
\input{dlab_macros}

\title{Integrating Machine-Generated Short Descriptions into the Wikipedia Android App: A Pilot Deployment of Descartes}
\author[1]{Marija Šakota\thanks{Corresponding author: marija.sakota@epfl.ch}}
\author[2]{Dmitry Brant}
\author[2]{Cooltey Feng}
\author[2]{Shay Nowick}
\author[2]{Amal Ramadan}
\author[2]{Robin Schoenbaechler}
\author[2]{Joseph Seddon}
\author[2]{Jazmin Tanner}
\author[2]{Isaac Johnson}
\author[1]{Robert West}

\affil[1]{EPFL}
\affil[2]{Wikimedia Foundation}
\date{}

\begin{document}
\maketitle

\begin{abstract}
\noindent
Short descriptions are a key part of the Wikipedia user experience, but their coverage remains uneven across languages and topics. In previous work, we introduced \textit{Descartes}, a multilingual model for generating short descriptions \cite{sakota2022descartes}. In this report, we present the results of a pilot deployment of Descartes in the Wikipedia Android app, where editors were offered suggestions based on outputs from Descartes while editing short descriptions. The experiment spanned 12 languages, with over 3,900 articles and 375 editors participating. Overall, 90\% of accepted Descartes descriptions were rated at least 3 out of 5 in quality, and their average ratings were comparable to human-written ones. Editors adopted machine suggestions both directly and with modifications, while the rate of reverts and reports remained low. The pilot also revealed practical considerations for deployment, including latency, language-specific gaps, and the need for safeguards around sensitive topics. These results indicate that Descartes's short descriptions can support editors in reducing content gaps, provided that technical, design, and community guardrails are in place. 
\end{abstract}

\section{Introduction}

Short descriptions play a central role in helping readers and editors quickly identify the topic of a Wikipedia article. For example, the short description of the article \textbf{\textit{Yukon}} is ``\textit{Territory of Canada}''. They appear prominently in search results, link previews, and mobile interfaces, and are often the first piece of information a user sees before deciding to read further. Despite their importance, coverage of short descriptions across Wikipedia editions remains uneven, and maintaining them at scale presents a significant challenge for volunteer editors. For many languages and article types, large gaps persist.

In earlier work, we introduced \textit{Descartes} (short for ``\textit{Desc}riber of \textit{art}icl\textit{es}''), a multilingual language model for generating concise and high-quality short descriptions directly from article content~\cite{sakota2022descartes}. Automatic evaluations and human assessments demonstrated that Descartes could approach the quality of human-written descriptions while generalizing across more than a dozen languages. These findings motivated the exploration of how such a system might be integrated into the editing workflows of Wikimedia projects, where human oversight and community trust are essential.

Encouraged by the first results, we initiated a pilot deployment of Descartes in the Wikipedia Android application. The goal of this pilot was to assess the practical value of Descartes short descriptions in a real-world setting: whether editors would use and adapt them, what benefits they might bring, and which modifications would be necessary for deployment. This report documents the design, implementation, and results of this experiment. Specifically, we describe the integration of Descartes into the Android short description tool, the methods used to evaluate its output, and the lessons learned from community feedback.  

The development and the results of this experiment were also published on a dedicated MediaWiki page,\footnote{\url{https://www.mediawiki.org/wiki/Wikimedia_Apps/Team/Android/Machine_Assisted_Article_Descriptions/Updates}} and are summarized here to inform future work on machine-assisted editing tools across Wikimedia projects.

\section{Experiment design and implementation}

\subsection{Short descriptions in the Wikipedia Android app and experiment outline}
Since 2017, editors on Wikipedia have been able to modify and add short descriptions in the Android application.\footnote{\url{https://www.mediawiki.org/wiki/Wikimedia_Apps/Team/Android/Machine_Assisted_Article_Descriptions\#Project_background}} In the following years, a dedicated tool for this purpose was made. Following our release of the Descartes model \cite{sakota2022descartes}, we reached out to the Wikimedia Foundation in mid-2022 with a proposal to improve this tool by providing editors with suggested descriptions using Descartes, subject to quality constraints and community approval. After initial exploration, the Android app team agreed to (1) instantiate an API for Descartes, (2) build the user interface (UI) and client-side logic, and (3) perform quality controls and embed guardrails.

The experiment was then carried out in two steps. In the first step, the experimental feature was released in the short-description tool, allowing editors to choose or modify them when editing short descriptions. In the second step, community graders rated the data collected in the first step on a scale from 1 to 5, providing insight on the quality of the short descriptions generated this way. They were also asked whether they would modify or revert the description. For the experiment, we used the ``Descartes[notype]'' model variant, the version of Descartes without semantic-type information. This was done to lower the amount of storage needed for the model, as semantic type embeddings can be several times bigger than the model itself.

\subsection{Experiment details}

\input{tables_figures/ui}

\xhdr{API}
The model is currently deployed on LiftWing,\footnote{\url{https://api.wikimedia.org/wiki/Lift_Wing_API/Reference/Get_article_descriptions}} Wikipedia's production machine learning serving platform. During the experiment, the model was deployed on Cloud VPS,\footnote{\url{https://wikitech.wikimedia.org/wiki/Portal:Cloud_VPS}} a more generic hosting platform. To generate a short description, the model requires the title of an article, the language in which it should be generated, and the number of beams to be used for decoding. We use beam search decoding \cite{sutskever2014sequencesequencelearningneural} that incrementally builds sequences by keeping only the top-$k$ highest-scoring partial sequences at each step, trading optimality for a tractable search. A higher number of beams means higher latency, but might also mean better performance.

\xhdr{User interface} After initial testing, the Android app team settled on a user interface for the experiment, integrating it with the existing short-description tool. Before starting the experiment, the editor could see a screen that explained what machine-generated descriptions are (see \Figref{fig:ui_pre}). Once the editor started editing, the interface allowed them to show two machine-generated descriptions (see \Figref{fig:ui1}, \Figref{fig:ui2}). If the editor selected one of them, the field was populated with the proposed description (see \Figref{fig:ui3}). They could then polish it by editing it, or remove it and write something of their own. The editor could also report the generated descriptions, in case they believed that it violated Wikipedia policies.

\xhdr{Guardrails}
After initial discussions about the model with the team and an exploration of its performance,\footnote{\url{https://public-paws.wmcloud.org/User:Isaac_(WMF)/Article\%20Descriptions/Harm_exploration.ipynb}} we identified the main risks and came up with the potential guardrails for bias and harm. They can be summarized as follows:
\begin{itemize}
    \item To prevent problematic text generation related to people, only editors with at least 50 edits saw recommendations for biographies of living people.
    \item To reduce low-quality suggestions, only two beams were used (instead of three).
    \item The accepted recommendations from the model were monitored throughout the experiment to identify if the experiment should be ended early.
\end{itemize}

Additionally, editors needed to have previously done at least three edits to be included in the experiment.

\xhdr{Latency}
During the experiment, the model was deployed on a server without any GPUs. In this setting, generating one short description can take up to 10 seconds, depending on the number of beams used during generation. To prevent problems with latency in the application, the team came up with a few solutions:
\begin{enumerate}
    \item Restricting the number of beams to a smaller number (two) in the application.
    \item Altering the tool such that the generations can be preloaded in the background.
\end{enumerate}

\xhdr{Languages}
The experiment was done in 12 languages (Arabic, Czech, German, English, Spanish, French, Gujarati, Hindi, Italian, Japanese, Russian, Turkish). For seven other languages (Finnish, Kazakh, Korean, Burmese, Dutch, Romanian, Vietnamese), editors had the opportunity to see machine-generated descriptions and select or modify them, but there were no community graders performing the quality check.

\section{Results}

\xhdr{Collection of grading data}
In the first part of the experiment, which lasted for a month and a half, 375 different editors were exposed to machine-generated descriptions for 3,968 different Wikipedia articles. Overall, 2,125 machine-generated descriptions were published by 256 editors. In order for their data to be processed for more advanced analysis, editors had to opt in for a more specific logging of actions, which leads to a smaller total number of edits in the next sections.
23.49\% (895) of machine-generated descriptions were accepted without changes and 14.49\% (552) were accepted with modifications. The rest of the machine-generated descriptions (2363 or 62.02\%) were rejected. Note that this number is higher than the real rejection rate because it also counts the situations in which editors did not reveal machine-generated description at all and hence could not accept it.

\input{tables_figures/grade_main}

\xhdr{Main results of the grading experiment}
In \Tabref{tab:grade}, we show the results of the grading experiment in different languages, as well as overall. For the machine-generated descriptions we only consider accepted descriptions, discarding the rejected ones. The results include the grade for the accepted machine-generated descriptions and human-written ones, as well as the recommendation if the feature should be enabled for each edition of Wikipedia, or if it should be enabled under certain conditions. The recommendations were made based on the average grade of machine-generated descriptions in comparison to the average grade of the human-generated ones for the specific Wikipedia edition. More conservative recommendations were made for the Wikipedia editions in languages for which we did not have many graders, and consequently, evaluated samples. Overall, 90\% of the accepted machine-generated descriptions received a grade of 3 or higher. Machine-generated descriptions that were not modified received an average grade of 4.2, while the modified ones received an average grade of 4.1.

\xhdr{Impact of editor's experience and beams}
In \Tabref{tab:editor_experience}, we show the results stratified by the experience of the editors. These results clearly indicate that more experienced editors are better at choosing which machine-generated descriptions to use. In \Tabref{tab:beams}, we show the results stratified by the beam number (i.e., order by which the descriptions were generated). We observe that the first-beam solution is chosen more often and produces slightly higher-quality descriptions. To control for ordering bias, suggestions were ordered randomly in the experiment.

\input{tables_figures/beams_editor_exp}

\xhdr{Reverts, modifications, and reports}
In \Tabref{tab:reverts_edits} we present the results from the second task community graders had to do: determine whether they would revert or edit the provided short description. Because only 20 machine-generated edits were reverted throughout this whole experiment, we could not compare the actual reverts to get statistically significant results. Instead, we estimate the revert and edit rates by directly asking the graders if they would revert or edit the descriptions. Reverts were defined as descriptions so bad that minor edits would not make any sense. For rewrites, patrollers\footnote{\url{https://meta.wikimedia.org/wiki/Meta:Patrollers}} would just be expected to edit the existing descriptions to improve them. In \Tabref{tab:reports}, we report the feedback that was received through reports of the machine-generated descriptions. Only 0.5\% editors used the report functionality.

\input{tables_figures/reverts_edits}

\input{tables_figures/reports_retention}

\xhdr{User retention}
In \Tabref{tab:retention}, we show the retention period for the control and treatment groups. In this setting, the control group was not exposed to machine-generated descriptions, while the treatment group was. Results indicate a slightly higher retention rate for users in the treatment group.

\section{Discussion and future directions}

The results of the experiment are overall promising, and in line with our expectations from the automatic and human evaluations that has been performed on Descartes prior to this experiment. Descartes's descriptions achieved a similar average grade to human-written ones, while at the same time increasing retention rate and lowering the amount of potential reverts. Based on the results, several modifications were made to prepare the model for usage in production:
\begin{itemize}
    \item \textbf{Restricted categories}: the biographies of living people are some of the most sensitive content Wikipedia editors work with \cite{10.1145/1940761.1940765}. Per the request of the Android app team, we recommend to exclude such content from the scope of our machine-generated descriptions in any full deployment, even though our assessments did not raise particular flags about this category. During the initial exploration, we explored other potentially problematic areas for biases, but did not find the need for further controls. In the future, we recommend monitoring the activity of the tool to identify any potentially problematic areas that would require special configuration.
    \item \textbf{Beam selection}: as the first beam consistently produced better results than the second one, only this beam should be presented to the editors.
    \item \textbf{Incorrect dates}: since this was a common error reported by the editors, descriptions with a date should only be shown if there is a support for that date in the text, to prevent model hallucinations.
    \item \textbf{Disambiguation pages}: since making a description for disambiguation pages is trivial (i.e., a canned description can be used), and Descartes sometimes recommended the canned description, these should be filtered.
    \item \textbf{Capitalization}: short descriptions in Wikipedia should start with a capital letter, so a simple scripted fix can be applied for such errors. 
\end{itemize}

For languages where the model is underperforming, the most effective way to improve is by increasing the number of short descriptions in that language in the training set. While there is no fixed retraining schedule as of now, this can be coordinated with communities that show interest in this feature. At this point, no further improvements of the model are planned, other than working on lowering the latency such that editors can see the suggestions more quickly.

Overall, the Android pilot demonstrates a promising path from a research model (Descartes) to an interactive editing aid. Results indicate that, although the model is robust, effective deployment requires careful UI design, safeguards, and community engagement. With improved infrastructure and continuous evaluation, Descartes descriptions may substantially reduce content gaps across languages, supporting both readers and editors.

\printbibliography

\end{document}

%% file: dlab_macros.tex
\usepackage[utf8]{inputenc}
\usepackage[T1]{fontenc}
\usepackage{hyphenat}
\usepackage{xspace}
\usepackage{amsmath}
\usepackage{amsfonts}
\usepackage{hyperref}
\usepackage{url}
\usepackage{booktabs}
\usepackage{multirow}
\usepackage{makecell}
\usepackage{caption}
\usepackage{minibox}
\usepackage{bbm}
\usepackage{graphicx}
\usepackage{balance}
\usepackage{mathtools}
\usepackage{color}
\usepackage{marvosym}
\usepackage{ifthen}
\usepackage{textcomp}
\usepackage{enumitem}
\usepackage{verbatim}
\usepackage{algorithm}
\usepackage{numprint}
\usepackage{balance}

\usepackage{amsthm}
\theoremstyle{plain}

\newcommand{\chatoDisplayMode}[1]{#1}

\definecolor{MyRed}{rgb}{0.6,0.0,0.0} 
\definecolor{MyBlack}{rgb}{0.1,0.1,0.1} 
\newcommand{\inred}[1]{{\color{MyRed}\sf\textbf{\textsc{#1}}}}
\newcommand{\frameit}[2]{
  \begin{center}
  {\color{MyRed}
  \framebox[.9\columnwidth][l]{
    \begin{minipage}{.85\columnwidth}
    \inred{#1}: {\sf\color{MyBlack}#2}
    \end{minipage}
  }\\
  }
  \end{center}
}

\newcommand{\note}[2][]{\chatoDisplayMode{\def\@tmpsig{#1}\frameit{{\Pointinghand} Note}{#2\ifx \@tmpsig \@empty \else \mbox{ --\em #1}\fi}}}
\newcommand{\todo}[2][]{\chatoDisplayMode{\def\@tmpsig{#1}\frameit{{\Writinghand} To-do}{#2\ifx \@tmpsig \@empty \else \mbox{ --\em #1}\fi}}}

\newcommand{\Tabref}[1]{Table~\ref{#1}}
\newcommand{\Figref}[1]{Fig.~\ref{#1}}

\newcommand{\xhdr}[1]{\vspace{1.7mm}\noindent{{\bf #1.}}}

\newcommand{\hide}[1]{}

\makeatletter
\newcommand{\iffont}[2]{\ifthenelse{\equal{\f@family}{#1}}{#2}{}}
\makeatother

%% file: tables_figures/ui.tex
\begin{figure}[htbp]
  \centering
  \begin{subfigure}[t]{0.24\textwidth}
    \includegraphics[width=\linewidth]{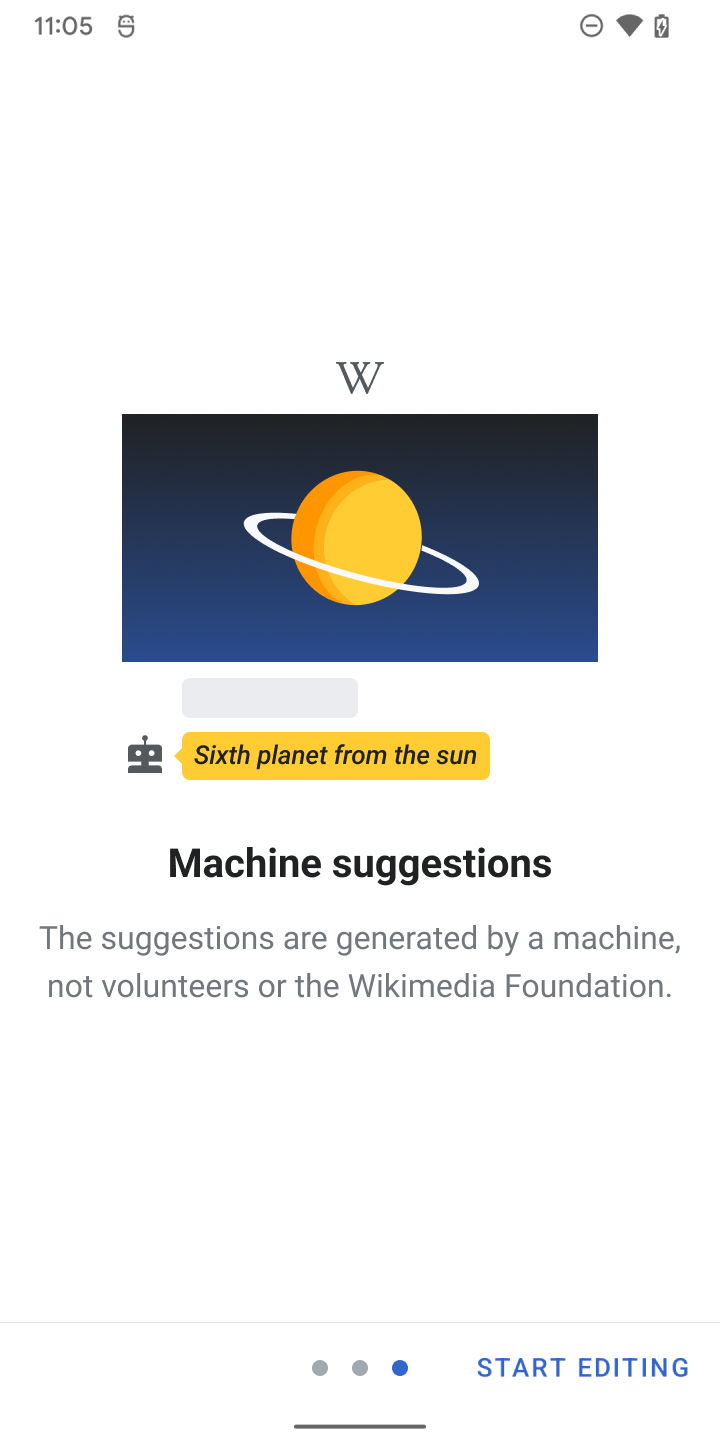}
    \caption{Onboarding screen \cite{JTannerImage_pre}.}
    \label{fig:ui_pre}
  \end{subfigure}\hfill
  \begin{subfigure}[t]{0.24\textwidth}
    \includegraphics[width=\linewidth]{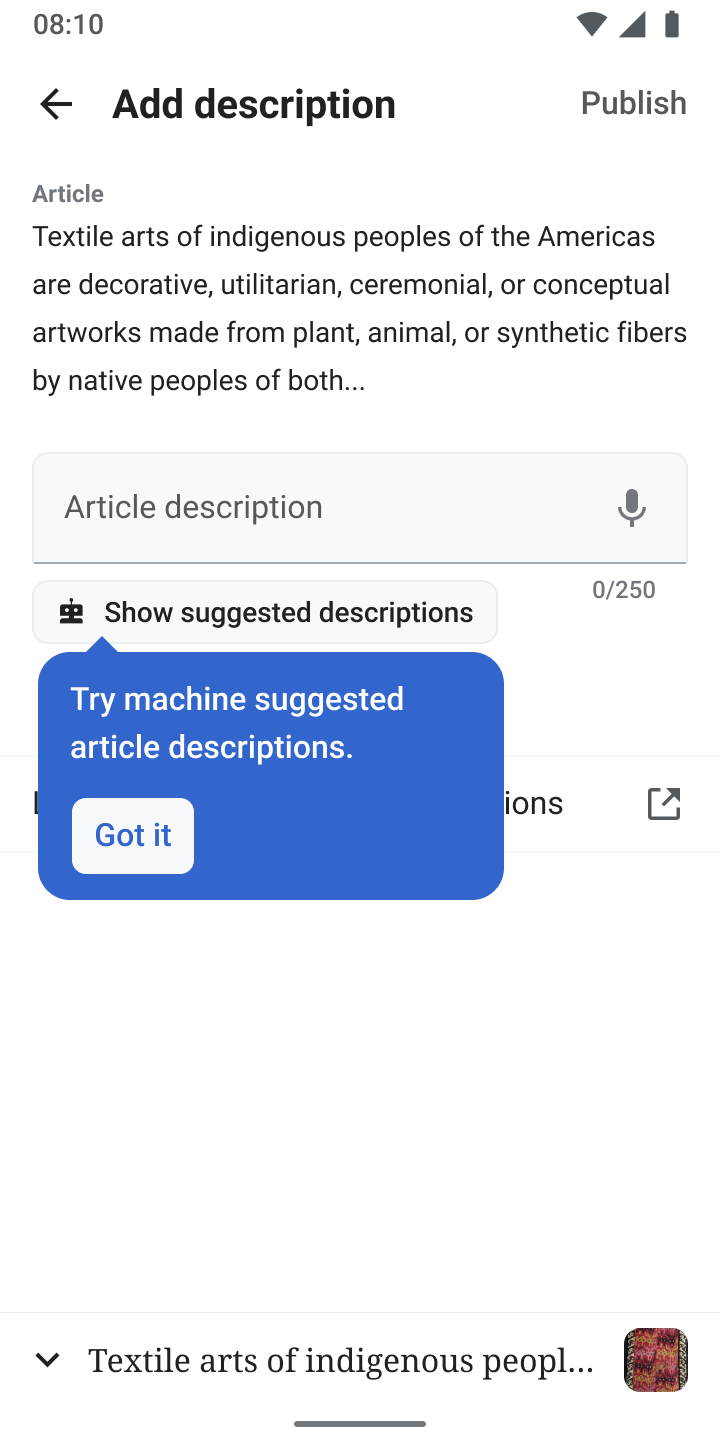}
    \caption{Editors can click to reveal the suggested descriptions \cite{JTannerImage_ui1}.}
    \label{fig:ui1}
  \end{subfigure}\hfill
  \begin{subfigure}[t]{0.24\textwidth}
    \includegraphics[width=\linewidth]{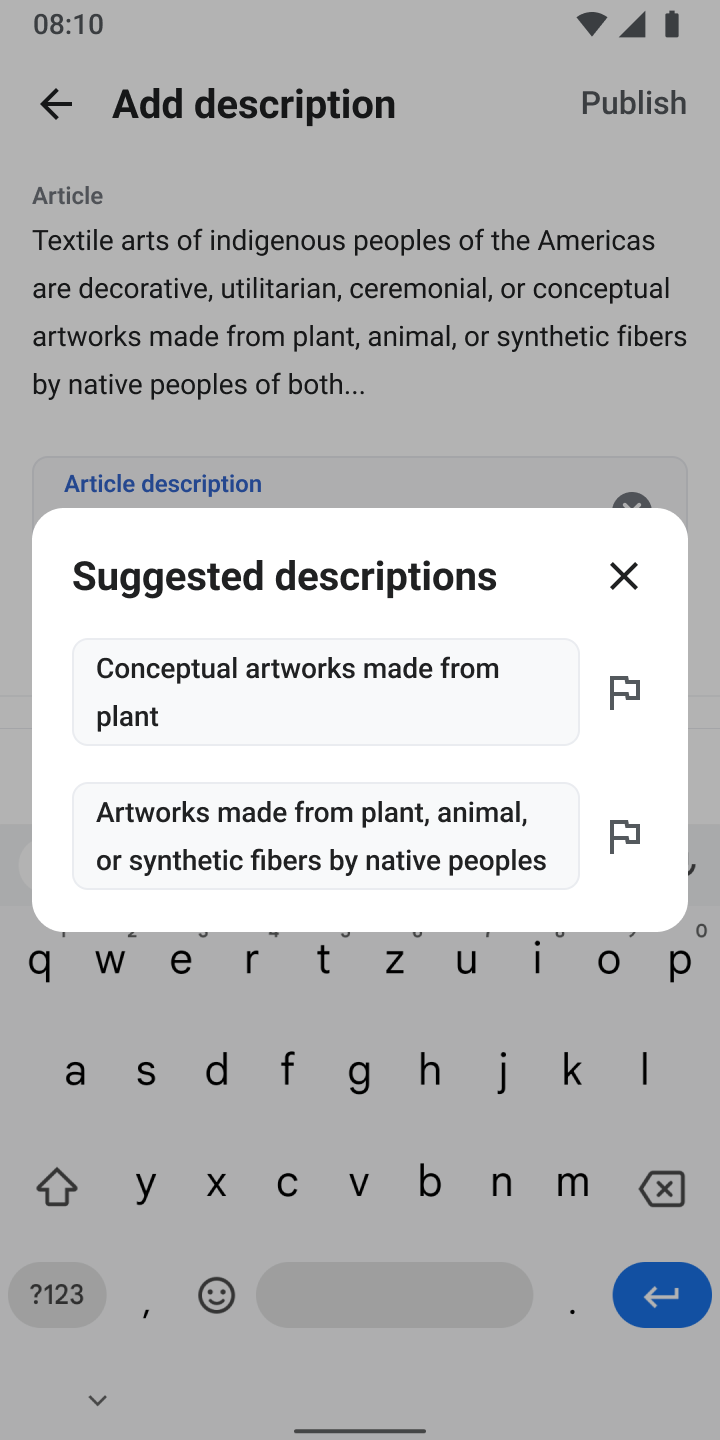}
    \caption{Screen with two suggested descriptions \cite{JTannerImage_ui2}.}
    \label{fig:ui2}
  \end{subfigure}\hfill
  \begin{subfigure}[t]{0.24\textwidth}
    \includegraphics[width=\linewidth]{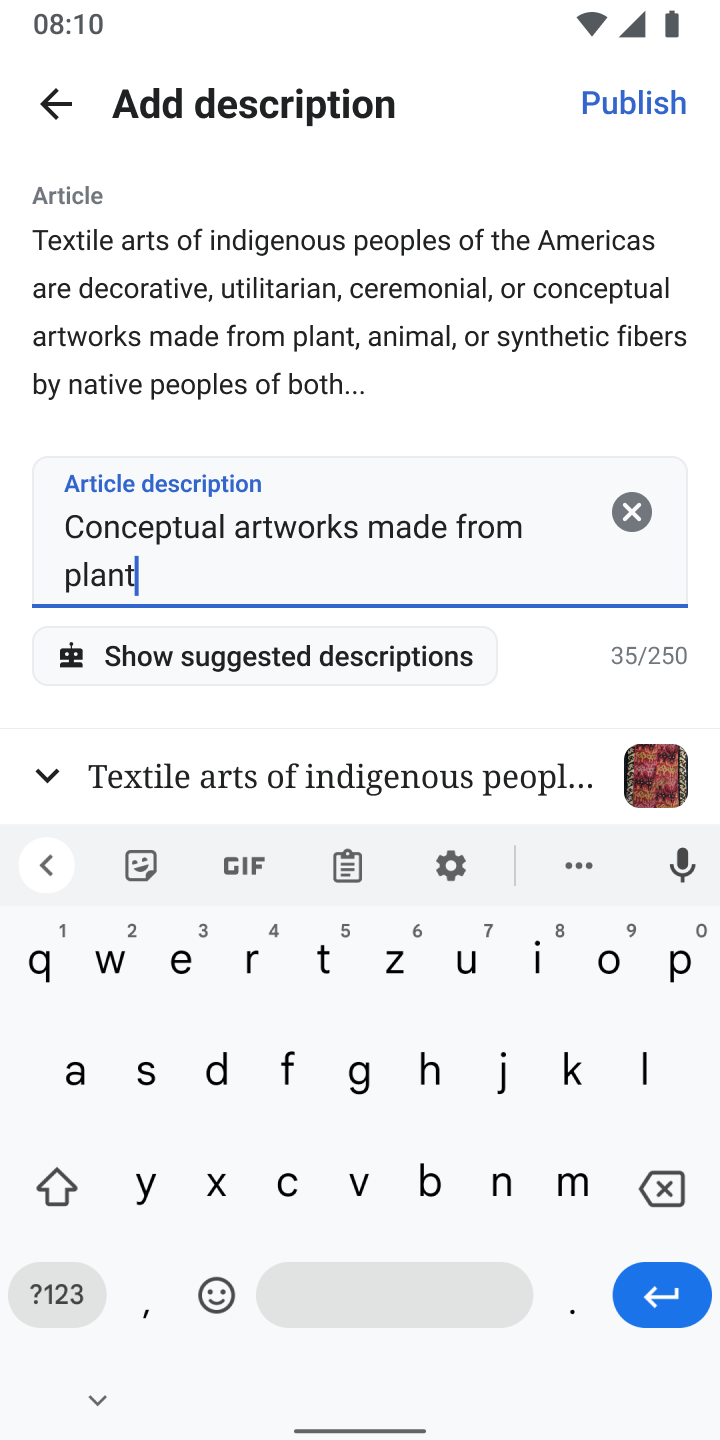}
    \caption{Editors can accept the selected suggestion as it is, or make modifications to it \cite{JTannerImage_ui3}.}
    \label{fig:ui3}
  \end{subfigure}

  \caption{Overview of the user interface.}
  \label{fig:four-images}
\end{figure}

%% file: tables_figures/grade_main.tex
\begin{table}[htbp]
\centering
\resizebox{\textwidth}{!}{
\setlength{\tabcolsep}{3pt}
\begin{tabular}{@{}lccc@{}}
\toprule
\textbf{Language}                   & \begin{tabular}[c]{@{}l@{}}\textbf{Accepted}\\ \textbf{machine-generated} \\ \textbf{desc. avg.\ grade}\end{tabular} & \begin{tabular}[c]{@{}l@{}}\textbf{Human-generated}\\
\textbf{desc. avg.\ grade}\end{tabular} & \begin{tabular}[c]{@{}l@{}}\textbf{Recommendation whether} \\ \textbf{to enable feature}\end{tabular} \\ \midrule \midrule
Arabic* (ar)                        & 2.8                                                                         & 2.1                                                                                              & No                                             \\
Czech (cs)                         & 4.5                                                                         &                                                                                                   & Yes                                            \\
German (de)                         & 3.9                                                                         & 4.1                                                                                             & 50+ edits required                             \\
English (en)                         & 4.0                                                                         & 4.5                                                                                             & 50+ edits required                             \\
Spanish (es)                         & 4.5                                                                         & 4.1                                                                                            & Yes                                            \\
French (fr)                         & 4.0                                                                         & 4.1                                                                                        & 50+ edits required                             \\
Gujarati* (gu)                        & 1.0                                                                         &                                                                                                  & No                                             \\
Hindi (hi)                         & 3.8                                                                         &                                                                                                   & 50+ edits required                             \\
Italian (it)                         & 4.2                                                                         & 4.4                                                                                              & 50+ edits required                             \\
Japanese (ja)                         & 4.0                                                                         & 4.5                                                                                              & 50+ edits required                             \\
Russian (ru)                         & 4.7                                                                         & 4.3                                                                                                & Yes                                            \\
Turkish (tr)                         & 3.8                                                                         & 3.4                                                                                               & Yes                                            \\ 
\hline
Overall & 4.1 & 4.2 & N/A \\
\bottomrule
\end{tabular}
}
\caption{Results of the community graders experiment. For Arabic and Gujarati (marked with *) there were not many suggestions collected to grade, which might have affected the score.}
\label{tab:grade}
\end{table}

%% file: tables_figures/beams_editor_exp.tex
\begin{table}[h!]
\centering
\begin{subtable}[t]{0.47\textwidth}
    \centering
    \resizebox{\textwidth}{!}{
    \setlength{\tabcolsep}{3pt}
    \begin{tabular}{@{}lcc@{}}
    \toprule
    Editor experience & Average desc. grade & Median desc. grade \\ \midrule
    \midrule
    Under 50 Edits    & 3.6                & 4                 \\
    Over 50 Edits     & 4.4                & 5                 \\ \bottomrule
    \end{tabular}
    }
    \caption{Grade for machine-generated descriptions depending on the experience of the editors.}
    \label{tab:editor_experience}
\end{subtable}\hfill
\begin{subtable}[t]{0.47\textwidth}
    \centering
    \resizebox{0.9\textwidth}{!}{
    \setlength{\tabcolsep}{3pt}
    \begin{tabular}{@{}lcc@{}}
    \toprule
    Beam selected & Average desc. grade & \% Distribution \\ \midrule
    \midrule
    1             & 4.2                & 64.7\%          \\
    2             & 4.0                & 35.3\%          \\ \bottomrule
    \end{tabular}
    }
    \caption{Performance of machine-generated options depending on the beam during generation. The options were randomly placed in the app to avoid positional bias.}
    \label{tab:beams}
\end{subtable}
\caption{Results stratified by editor's experience and beam.}
\label{tab:beams_editor_exp}
\end{table}

%% file: tables_figures/reverts_edits.tex
\begin{table}[htbp]
\centering
\resizebox{\textwidth}{!}{
\setlength{\tabcolsep}{3pt}
\begin{tabular}{@{}lcc@{}}
\toprule
Graded descriptions:                                                     & \% desc. would revert & \% desc. would rewrite \\ \midrule
\midrule
Editor accepted model's suggestion                                        & 2.3\%                 & 25.0\%                 \\
Editor saw suggestion but wrote out their own description instead & 5.7\%                 & 38.4\%                 \\
Human edit no exposure to suggestion                              & 15.0\%                & 25.8\%                 \\ \bottomrule
\end{tabular}
}
\caption{Percentage of edits that community graders would either revert or rewrite. The first row corresponds to the accepted machine-generated descriptions, the second to the human-written ones when the machine-generated description was available but rejected (either with or without revealing it), and the third row corresponds to the human-written descriptions, when no machine-generated description was available.}
\label{tab:reverts_edits}
\end{table}

%% file: tables_figures/reports_retention.tex
\begin{table}[htbp]
\centering
\begin{subtable}[t]{0.47\textwidth}
  \centering
  \adjustbox{valign=t,width=\linewidth}{%
    \setlength{\tabcolsep}{3pt}%
    \begin{tabular}{@{}lc@{}}
      \toprule
      Feedback/Response        & \% Distribution of feedback \\ \midrule
      \midrule
      Not enough info          & 43\%                        \\
      Inappropriate suggestion & 21\%                        \\
      Incorrect dates          & 14\%                        \\
      Cannot see description   & 7\%                         \\
      ``Unnecessary hook''       & 7\%                         \\
      Faulty spelling          & 7\%                         \\ \bottomrule
    \end{tabular}
  }
  \caption{Distribution of responses through report function.}
  \label{tab:reports}
\end{subtable}\hfill
\begin{subtable}[t]{0.47\textwidth}
  \centering
  \adjustbox{valign=t,width=\linewidth}{%
    \setlength{\tabcolsep}{3pt}%
    \begin{tabular}{@{}lcc@{}}
      \toprule
      Retention Period            & Control & Treatment \\ \midrule
      \midrule
      1-day average return rate:  & 35.4\% & 34.9\% \\
      3-day average return rate:  & 29.5\% & 30.3\% \\
      7-day average return rate:  & 22.6\% & 24.1\% \\
      14-day average return rate: & 14.7\% & 15.8\% \\ \bottomrule
    \end{tabular}
  }
  \caption{Retention period for treatment and control group. Treatment group (that was exposed to machine-generated descriptions) has slightly higher retention rate than control (not exposed to machine-generated descriptions).}
  \label{tab:retention}
\end{subtable}
\caption{Statistics on report function and retention period.}
\label{tab:both}
\end{table}